\pdfoutput=1

\documentclass[11pt]{article}

\usepackage[]{acl}

\usepackage{times}
\usepackage{latexsym}

\usepackage[T1]{fontenc}

\usepackage[utf8]{inputenc}

\usepackage{microtype}

\usepackage{url}
\usepackage{amsmath,amssymb}
\DeclareMathOperator{\E}{\mathbb{E}}
\usepackage{amsfonts}
\usepackage{helvet}
\usepackage{courier}
\usepackage{url}
\usepackage{float,color}
\usepackage{times}
\usepackage{algorithm}
\usepackage{algorithmic}
\usepackage{epsfig}
\usepackage{stfloats}
\usepackage{url}
\usepackage{diagbox}
\usepackage{subfigure}
\usepackage{epstopdf}
\usepackage{multicol}
\usepackage{dsfont,amsfonts,amssymb,amsmath,color}
\usepackage{multirow}
\usepackage{booktabs}
\def\0{{\bf 0}}
\def\1{{\bf 1}}

\def\LM{{\mathcal L}}
\def\RM{{\mathcal R}}

\def\NM{{\mathcal N}}

\def\SM{{\mathcal S}}

\title{Weakly Supervised Text Classification using Supervision Signals \\
from a Language Model}

\author{Ziqian Zeng$^{1,2}$, Weimin Ni$^{1}$, Tianqing Fang$^{2}$, Xiang Li$^3$, Xinran Zhao$^4$, and Yangqiu Song$^2$ \\
  $^1$Shien-Ming Wu School of Intelligent Engineering, South China University of Technology, China\\
  $^2$Department of CSE, HKUST, Hong Kong, China\\
  $^3$Department of Computer Science, University of Illinois, Urbana Champaign, US \\
  $^4$Computer Science Department, Stanford University, US \\
  {\tt zqzeng@scut.edu.cn},
  {\tt mewmn@mail.scut.edu.cn}, {\tt xiangl12@illinois.edu} \\
  {\tt xzhaoar@stanford.edu}, {\tt \{tfangaa, yqsong\}@cse.ust.hk } \\
}

\begin{document}
\maketitle
\begin{abstract}
Solving text classification in a weakly supervised manner is important for real-world applications where human annotations are scarce. 
In this paper, we propose to query a masked language model with cloze style prompts to obtain supervision signals. 
We design a prompt which combines the document itself and ``this article is talking about {\tt[MASK]}.'' 
A masked language model can generate words for the {\tt[MASK]} token. 
The generated words which summarize the content of a document can be utilized as supervision signals. 
We propose a latent variable model to learn a word distribution learner which associates generated words to pre-defined categories and a document classifier simultaneously without using any annotated data. 
Evaluation on three datasets, AGNews, 20Newsgroups, and UCINews, shows that our method can outperform baselines by 2\%, 4\%, and 3\%. 
\end{abstract}
\section{Introduction}
Text classification is a fundamental task in Natural Language Processing (NLP) with diverse real-world applications such as identifying relevant documents of a case in legal proceedings \cite{roitblat2010document}, and classifying victim's requests (e.g., food, shelter, and medical aids) on social media platforms during earthquakes \cite{caragea2011classifying}. 
Current state-of-the-art text classification methods \cite{zhang2015character,zhou2016text,johnson2017deep} still need a large number of annotated data. 
However, in the real world, naturally annotated data are rare and human annotations are expensive. 
Solving the text classification task without using annotated data but exploiting inexpensive supervision signals is worth investigation. 

In the weakly supervised setting, any annotated document is not accessible, but inexpensive supervision signals such as label surface names or keywords can be used. 
Existing weakly supervised text classification methods \cite{meng2018weakly,meng2019weakly,mekala2020contextualized,meng2020text} first used seed keywords to retrieve more keywords, and then created pseudo labels for documents and then train a model in a ``standard'' supervised learning manner. 
In previous work, supervision signals are restricted to a small set of keywords from documents contents.

Recent work shows that prompts can probe knowledge from PLMs \cite{devlin2018bert,radford2019language} and the knowledge can provide supervision signals to solve different NLP tasks including relation extraction \cite{shin2020autoprompt}, question answering \cite{petroni2020context}, and summarization \cite{radford2019language}. 
For example, \cite{petroni2019language} solved the knowledge base completion task by querying an MLM with a prompt ``Alan Turing was born in {\tt[MASK]}.'' 
Using prompts to generate supervision signals for text classification is worth exploring.

\begin{figure*}[h]
\centering
\includegraphics[width=\textwidth]{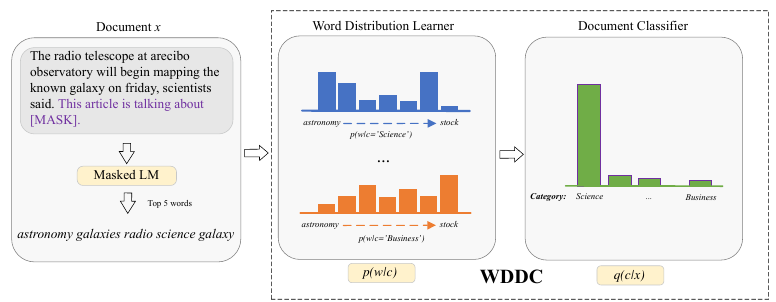}

\caption{
We combine a document and a cloze style sentence ``This article is talking about {\tt[MASK]}'' to query a masked LM. 
It generates a set of words for the {\tt[MASK]} token. 
These words are likely to summarize the topic of a document. 
After obtaining words such as ``astronomy'' and ``galaxies'', human beings can easily infer that this article is talking \textit{science} rather than \textit{business} because we know these words are frequently used in \textit{science} topic. 
Word distributions given pre-defined categories bridge supervision signals (generated words) and our goal (the category of a document). 
The proposed model (WDDC) can learn word distributions given pre-defined categories and a document classifier simultaneously.
} 
\label{fig:example}
\end{figure*}


We propose to query an MLM with a prompt which combines the document itself and ``this article is talking about {\tt[MASK]}.'', and use generated words for the {\tt[MASK]} token as supervision signals. 
For example, in Figure \ref{fig:example}, given a prompt ``The radio telescope at arecibo observatory will begin mapping the known galaxy on friday, scientists said. This article is talking about {\tt[MASK]}.'', an MLM predicts ``astronomy'', ``galaxies'', ``radio'', ``science'', and ``galaxy'' for the {\tt[MASK]} token.
These words summarize the topic of the document. 
Hence, they can be used as supervision signals. 
Besides generating signal words, an intuitive approach to obtain supervision signals is by extracting important words from documents. 
We will compare two types of supervision signals. 

After obtaining signal words, we need to associate these words to pre-defined categories. 
We propose a latent variable model (\textbf{WDDC}) to learn a \textbf{W}ord \textbf{D}istribution learner and a \textbf{D}ocument \textbf {C}lassifier simultaneously without using any annotated data. 
A word distribution learner aims to learn a probability of a generated word $w$ given a category $c$, i.e., $p(w|c)$. 
A document classifier aims to learn a probability of a category $c$ given a document $x$, i.e., $p(c|x)$. 
These two goals could be optimized simultaneously via maximizing the log-likelihood of generated words by introducing the category as a latent variable. 
In our latent variable model, a word distribution learner and a document classifier can be parameterized by any neural network.


Our contributions are summarized as follows,

$\bullet$ We propose to query an MLM using a prompt which combines the document and a cloze style sentence ``this article is talking about {\tt[MASK]}''. We use generated words for the {\tt[MASK]} token as supervision signals in the weakly supervised text classification task. 

$\bullet$ We propose a latent variable model (WDDC) to learn a word distribution learner which associates generated words to pre-defined categories and a document classifier simultaneously without using any annotated data.

$\bullet$ The experimental results show that the proposed method WDDC can outperform other weakly supervised baselines in three datasets. 

The code is available at \url{https://github.com/HKUST-KnowComp/WDDC}.
\section{Related Work}
In this section, we review the related work on querying an MLM with prompts, weakly supervised text classification, zero-shot text classification, and variational methods.

\paragraph{Querying an MLM with Prompts.}
Querying an MLM with cloze style prompts provides a new direction to solve some NLP tasks in an unsupervised manner. \cite{petroni2019language} queried an MLM using manually designed prompts to solve a knowledge base completion task. 
For example, in order to complete the missing entity {\tt X} in (Alan Turing, born in, {\tt X}), they designed a prompt ``Alan Turing was born in {\tt [MASK]}'' to query an MLM. 
The generated word for the {\tt [MASK]} token can be directly used to complete the missing fact. 
By querying language models, some NLP tasks such as relation extraction \cite{shin2020autoprompt}, question answering \cite{radford2019language,petroni2020context}, summarization \cite{radford2019language} could be solved in an unsupervised manner. 
However, not all NLP tasks can directly use generated words from an MLM in downstream tasks. 
Some tasks such as sentiment analysis and textual entailment \cite{shin2020autoprompt} need more steps for inference. 
For example, in the sentiment analysis task, \cite{shin2020autoprompt} used annotated data to train a classifier that links generated words to pre-defined categories. 
Our work does not require annotated data for inference.

\paragraph{Weakly Supervised Text Classification.}
In the weakly supervised text classification task, any labeled documents are not allowed, but label surface names or limited word-level descriptions of each category can be used. 
Dataless \cite{chang2008importance,song2014on} used Explicit Semantic Analysis (ESA) vectors \cite{gabrilovich2007computing} to represent label name and documents. 
Predictions are based on the label-document similarity. 
Recently, \cite{meng2018weakly,meng2019weakly,mekala2020contextualized,meng2020text,schick2021exploiting,schick2021small,zhang2022prboost} trained neural text classifiers in an weakly supervised manner.  
They generated pseudo labels for documents to pre-train a neural classifier and then performed self-training on unlabeled data for model refinement.
LOTClass \cite{meng2020text} is relevant to our work because they also used pre-trained language models. 
They used a LM to retrieve a set of semantically correlated words for each class, and then fine-tuned the LM to predict these words. 
Finally, they performed self-training on unlabeled data. 
Our work is different from LOTClass because we obtain supervision signals by querying an MLM with cloze style prompts and we propose a latent variable model to learn document classifier rather than using the self-training procedure. 
PRBOOST \cite{zhang2022prboost} is also relevant to our work because they also use prompts to generate weak labels. 
PRBOOST first generated rules by using a small amount of labeled data, then asked human annotators to select high-quality rules to generate week labels. Finally, they trained a new model in a self-training manner. 
Our work is different from PRBOOST because our method associates predicted words with labels in an unsupervised manner while PRBOOST maps prompting based rules to labels by involving human feedback.  

\paragraph{Zero-Shot Text Classification.} 
In zero-shot learning settings, the classes covered by training instances and the classes we aim to classify are disjoint. 
Zero-shot learning text classification methods \cite{xia2018zero, kavuluru2018riloff, zhang2019integraing, liu2019reconstructing} generalized seen classes to unseen classes by learning semantic relationships between classes and documents via embeddings or semantic knowledge sources. 
However, zero-shot learning still requires annotated data for the seen classes training. 
We cannot apply zero-shot learning methods to weakly supervised settings where no annotated document is available. 

\paragraph{Variational Methods.}

Variational autoencoders \cite{KingmaW13,rezende2014stochastic} consists of an encoder and a decoder. 
The encoder estimates posterior probabilities and the decoder estimates the reconstruction likelihood given a latent variable. The objective function is to maximize the reconstruction likelihood of the observed variable.
The latent variable in VAEs is continuous variable. 
Recently, many research works  \cite{titov2014unsupervised,marcheggiani2016discrete,vsuster2016bilingual,zhang2018variational,chen2018variational,zeng2019variational,liang2019relation} use VAEs to solve different NLP tasks such as relation discovery, question answering, sentiment classification, etc. 
In above works, the latent variables are discrete variables.
For example, \cite{marcheggiani2016discrete} aimed to solve unsupervised open-domain relation discovery. 
The objective function is to reconstruct the likelihood of two entities. 
They introduced relation as the latent variable.
The encoder is a relation classifier, which predicts a semantic relation between two entities. The decoder reconstructs entities given the predicted relation.   
Our method is also based on VAEs with a discrete latent variable but the estimated probabilities and the objective function are different.

\section{Methodology}
In this section, we first introduce how to obtain supervision signals from an MLM and document itself, and then we introduce a latent variable model to learn a word distribution learner and a document classifier simultaneously. 
\begin{table*}[ht!]
  \centering
  \caption{Signal words from an MLM and from the document(Doc).}
  \begin{tabular}{l|l|l}
  \toprule
  Text & Label &Signal Words \\
  \midrule[1pt]
  The world 's top two players  &  Sports & MLM: tennis, thailand, federer, seeds, wimbledon  \\
  roger federer and andy roddick &  &\\
  reached the semifinals friday  & & Doc:    world, players, federer, andy, roddick, semifinals, \\ 
  at the thailand open. & & friday, thailand, open\\ 
  \midrule
  These circuits abound in most & Science &  MLM: circuits, computers, electronics, computing, graphs \\ 
  electronic project books. &  & \\
  It has LED indicators also. & & Doc:   circuits, project, books, LED, indicators \\ 
  \midrule
  Scientists discover a genetic  & Health &  MLM: suicide, genetics, cancer, hiv, health \\
  indicator that could help & &\\
  prevent suicides. && Doc:   scientists, indicator, suicides \\
  \bottomrule
  \end{tabular}
  \label{tab:signal_example}
\end{table*}
\subsection{Supervision Signals}
\subsubsection{Signal Words}
Given a document, our goal is to obtain topic relevant words which are used as supervision signals. 
To achieve this, we append a cloze style sentence to the document at the end as a prompt. 
A prompt is designed as ``{\tt [CLS]} $+$ document $+$ This article is talking about {\tt [MASK]}. $+$ {\tt[SEP]}.''
The {\tt [MASK]} token serves as a placeholder for a topic relevant word which can summarize the document. 
It mimics the reading comprehension task which is using a word to summarize the content of a document. 
We select top $k$ generated words as supervision signals.

Instead of generating signal words, a natural way to obtain supervision signals is by extracting words from the document.
To achieve this, we extract all nouns and proper nouns in the document using part-of-speech tagger \cite{toutanova2003feature}. 
Since most of the generated words from an MLM are nouns and proper nouns, so we only extract words with two types of part-of-speech.

Table \ref{tab:signal_example} shows top $5$ predictions from an MLM (BERT \cite{devlin2018bert}) given prompts and extracted nouns and proper nouns from documents. 
For the first document, an MLM can infer that it is talking about a tennis match although ``tennis'' does not appear in the document. 
It also generates some relevant words such as `` wimbledon.'' 
In this case, the MLM is better than extraction. 
For the second document, the first word from the MLM precisely summarizes the document. 
However, the MLM also generates a few words which are related to \textit{computer}. 
Unfortunately \textit{computer} is also a category in this dataset. 
Compared to the MLM, the extracting way is safer in this case.  
For the third document, an MLM generates ``health'' which is an exact match of the label surface name although ``cancer'' and ``hiv'' are not faithful to the original document. We will evaluate generation and extraction methods in the experiment. 

\subsubsection{Remove Non-discriminated Words}
 Words generated from an MLM are not always category discriminated. Non-discriminated words can harm the performance of inference. 
The intuition of removing non-discriminated words is that if some words appear in different categories with similar frequency, then it is possible that these words are not category-discriminated. 
Since we cannot access labels, the label in the following computation means the pseudo label. 
The pseudo label generation process is shown in section \ref{sec:pseudo}. 
Inspired by category-indicative measurement,  \cite{mekala2020contextualized}, we define category-indicative index: 
\begin{equation}
    CII(c_i,w) = \frac{f(c_i,w)}{f(c_i)},
    \label{eq:cci}
\end{equation}
where $f(c_i,w)$ is the number of occurrences of the signal word $w$ in the documents which are labeled as $c_i$, and $f(c_i)$ is the total number of occurrences of all signal words in documents which are labeled as $c_i$. 

We define category-indicative ratio as,
\begin{equation}
    CIR(w) = \frac{CII(c_i,w)}{CII(c_j,w)},
    \label{eq:ccr}
\end{equation}
where $CII(c_i,w)$ is the maximum value among all categories,
$CII(c_j,w)$ is the second maximum value all categories. 
Larger value of $CIR(w)$ indicates $w$ is more discriminated. 
If $CII(c_j,w)$ is equal to 0, we will assign a large value to $CIR(w)$.
If $CIR(w) < t$, we consider $w$ is not discriminated and we remove $w$ from signal words set. 

\subsubsection{Pseudo Label Generation}
\label{sec:pseudo}
We assign pseudo labels to data based on label-word similarity. 
We represent a word using static representation which is introduced by \cite{mekala2020contextualized}. 
Given a word $w$, static representation $\SM\RM(w)$ is computed by averaging the contextualized embeddings of all its occurrences in the corpus. 
The label-word similarity is the cosine similarity between the static representation of the label surface name and the static representation of signal words. 
If the label surface name or the supervision signal contains more than one word, we take the average of the static representations of all words. 
We assign a sample with the pseudo label which yields the maximum similarity value among all classes. 
And the similarity value should be greater than a threshold $\gamma$. 
Setting a threshold can result in more accurate pseudo label assignments although the size of pseudo labeled data will shrink.

To summarize, there are three steps to obtain clean signal words:
(1) Obtain signal words from an MLM or a document. (2) Generate pseudo labels. (3) Remove signal words which have low category-indicative ratio values.

\subsection{Model Training}
After getting clean signal words, we then propose a latent variable model to learn a word distribution learner and a document classifier simultaneously.

Since there is no annotated data available, in order to best explain the observed data, i.e., signal words, the objective of our model is to maximize the log-likelihood of signal words. 
The ultimate goal is to identify the category of a document, hence, we introduce a latent variable $C$ representing the category, into the objective function. 
Further, by applying Jensen's inequality \cite{jensen1906fonctions}, we can derive an evidence lower bound (ELBO) of the log-likelihood. 
We define the objective function as follows,
\begin{align}
\LM_o & = \sum_{x \in X} \sum_{w_r \in \RM_{x}} \log p(w_r) \nonumber \\
& = \sum_{x \in X} \sum_{w_r \in \RM_{x}} \log \sum_{c}p(w_r , c) \nonumber\\
& = \sum_{x \in X} \sum_{w_r \in \RM_{x}}  \log \sum_{c} q(c|x) \Big[ \frac{p(w_r , c )}{q(c|x)} \Big] \nonumber \\
& \geq \sum_{x \in X} \sum_{w_r \in \RM_{x}}  \sum_{c} q(c|x) \Big[ \log \frac{p(w_r , c)}{q(c|x)} \Big] \nonumber \\
& =  \sum_{x \in X} \sum_{w_r \in \RM_{x}}\E_{q(C|x)} \big[\log p(w_r|c)p(c) \big] \nonumber \\ 
&  - \sum_{x \in X} \sum_{w_r \in \RM_{x}} \E_{q(C|x)}\big[ \log q(c|x)\big] , \label{eq:objective_one}
\end{align}

where $x$ is a document, $X$ is a set of documents,  $\RM_x$ is the set of signal words of document $x$, $w_r$ is a signal word, 
$C$ is a discrete random variable representing the category of a document, 
$c$ is a possible value of variable $C$. 
For example, $c$ can be \textit{science} or \textit{business}.

There are three probabilities in the Eq. (\ref{eq:objective_one}). 
$q(c|x)$ is the document classifier which is our ultimate goal.
$p(w_r|c)$ is the word distribution learner which estimates the probability distribution of all signal words given a possible value $c$. 
We use neural networks to parameterize $p(w_r|c)$ and $q(c|x)$.  
$p(C)$ is a prior probability distribution. 
Since there are no annotated data available, we cannot estimate $p(C)$.  
Hence we assume it is a uniform distribution, and $p(c)$ becomes a constant. 

\begin{table*}
  \centering
  \caption{Statistics and label surface names in AGNews, 20Newsgroup, and UCINews.}
  \begin{tabular}{l|l|l|l|l|l}
  \toprule
  Datasets & \# Train & \# Dev & \# Test & \# Class & Label Surface Names  \\
\midrule[1pt]
  AGNews & 108,000 & 12,000 & 7,600 & 4 & politics, sports, business, technology \\
  \midrule
  & & & & & computer graphics, \\
  & & & & & sports car, \\
  20Newsgroup & 14,609 & 1,825 & 1,825 & 6 & science electronics encryption health aerospace,  \\
  & & & & & politics gun homosexuality, \\
  & & & & & religion atheist christianity, \\
  & & & & & sale \\
  \midrule
  UCINews  & 26,008 & 2,560 & 27,556 & 4 & entertainment, technology, business, health \\
  \bottomrule
  \end{tabular}
  \label{tab:stat}
\end{table*}

\subsubsection{Word Distribution Learner} \label{sec:word_dist}
The word distribution learner aims to estimate the probability of a signal word $w_r$ given a possible value of category $c$. It is defined as follows,
\begin{equation}
p(w_r |c) = \frac{\exp \big( \mathbf{v}_{c}^T \mathbf{w}_r \big)}{ \sum_{{w_{r'} }}{ \exp \big( \mathbf{v}_{c}^T \mathbf{w}_{r'} \big) } } \;,
\label{eq:word_dist}
\end{equation}
where $\mathbf{v}_{c}$ is a trainable vector associated with $c$ and $\mathbf{w}_r$ is the trainable word embedding of signal word $w_r$. 
The intuition is that if a word (e.g., ``scientist'') appears frequently under the \textit{science} category, the corresponding inner-product value is high, otherwise it is low.

Eq. (\ref{eq:word_dist}) requires the summation over all signal words. 
Since the size of the word vocabulary can be large, we use the negative sampling technique \cite{mikolov2013distributed} to approximate Eq. (\ref{eq:word_dist}).
Specifically, we approximates $\log p(w_r|c)$ as follows,
\begin{equation}
\log \sigma \big( \mathbf{v}_{c}^T \mathbf{w}_r \big) + \sum_{ w_r' \in \NM } \log\big( 1 - \sigma \big(\mathbf{v}_{c}^T \mathbf{w}_r' \big) \big) \; \label{eq:word_dist_appro},
\end{equation}
where $w_r'$ is a negative sample in the vocabulary, $\NM$ is the set of negative samples and $\sigma(\cdot)$ is the sigmoid function.

The objective function with an approximated word distribution learner is defined as follows,

\begin{align} 
\LM & = \sum_{x \in X} \sum_{w_r \in \RM_{x}}  \E_{q(C|x)} \big[ \log \sigma \big( \mathbf{v}_{c}^T \mathbf{w}_r \big) \nonumber \\
\quad \quad & + \sum_{ w_r' \in \NM } \log \big( 1 - \sigma \big( \mathbf{v}_{c}^T \mathbf{w}_r' \big) \big) + \log p(c) \big] \nonumber \\
\quad \quad & - \E_{q(C|x)}\big[ \log q(c|x)\big]  \label{eq:la}.
\end{align}

\subsubsection{Document Classifier}
\label{sec:doc_classifier}
Most existing deep neural models (DNN) can be used to parameterize $q(C|x)$. 
As long as the input of DNNs is a document, and the output is a probability distribution of category $C$.
Since models which involve latent variables are difficult to optimize, we give a good initialization of the document classifier. 
We pre-train the document classifier using pseudo labeled data to initialize it. 

\section{Experiments}
In this section, we show the empirical performance of our method on the text classification task. 
\subsection{Datasets}

\begin{table}[t!]
  \centering
  \caption{Vocabulary size of signal words that are generated from an MLM and that are extracted from the document (Doc) after removing non-discriminated words. }
  \begin{tabular}{l|l|l}
  \toprule
  {Dataset}& MLM & Doc\\
  \toprule
AGNews  &  724 & 584 \\

20Newsgroup & 1,037 & 413 \\ 

UCINews & 584 & 442 \\
  \bottomrule
  \end{tabular}
  \label{tab:vocab_size}
\end{table}

We evaluate all methods on three datasets. 

(1) \textbf{AGNews} consists of news articles. It is constructed by \cite{zhang2015character}, which has been gathered from more than 2000 news sources in more than one year of activity. 

(2) \textbf{20Newsgroup} comprises around 18,000 posts. It is originally collected by \cite{lang1995newsweeder}. We perform text classification on coarse-grained topics. 
It is an unbalanced dataset.

(3) \textbf{UCINews} consists of news pages collected from a web aggregator. It is maintained by \cite{dua2017uci}. 

Table \ref{tab:stat} provides statistics and label surface names of three datasets. 
In 20Newgroups, we expand label surface names by combining fine-grained label surface names under the same coarse-grained category.

Table \ref{tab:vocab_size} shows the vocabulary size of signal words that are generated from an MLM and that extracted from the document (Doc) after removing non-discriminated words. 


\begin{table*}[ht!]
  \centering
  \caption{Micro F1 and macro F1 scores of all methods on AGNews, 20Newsgroup, and UCINews.}
    \begin{tabular}{l|cc|cc|cc}
    \toprule
     \backslashbox[48mm]{Methods}{Datasets}&
     \multicolumn{2}{c}{AGNews} & \multicolumn{2}{|c}{20Newsgroup} &
     \multicolumn{2}{|c}{UCINews} \\
     & Micro & Macro & Micro & Macro & Micro & Macro \\
    \midrule 
      Dataless \cite{chang2008importance} &  0.6855 & 0.6844  &  0.5000 & 0.4700 &  0.6248 & 0.6253  \\
      Label-Word Similarity & 0.7917 & 0.7884 & 0.7310 & 0.6390 & 0.6447 & 0.6390 \\
      {Pseudo-CNN} & 0.8265 & 0.8237 & 0.7973 & 0.6825 & 0.7598 & 0.7632 \\
      {Pseudo-BERT} & 0.8249 & 0.8219 & 0.8153 & 0.6896 & 0.7824 & 0.7820 \\
      WeSTClass \cite{meng2018weakly} &  0.8279 & 0.8268  &  0.5300 & 0.4300  &  0.6983 & 0.6999 \\
      LOTClass \cite{meng2020text} &  0.8659 & 0.8656 &  0.6121 & 0.5586  &  0.7320 & 0.7236  \\
      ConWea \cite{mekala2020contextualized} &  0.7443 & 0.7401 & 0.6200 & 0.5700 & 0.3293 & 0.3269\\
      X-Class \cite{wang2021xclass} & 0.8574 & 0.8566 & 0.6515 & 0.6316 & 0.6885 & 0.6962 \\
      \midrule
      WDDC-MLM & \textbf{0.8826} & \textbf{0.8825} & 0.8121 & 0.6882 & \textbf{0.8150} & \textbf{0.8134}   \\
      
      WDDC-Doc & 0.8668 & 0.8657 & \textbf{0.8570} & \textbf{0.8250} & 0.7814 & 0.7772 \\
      
      \midrule  
      CNN \cite{kim2014convolutional} &  0.9025 & 0.9025  & 0.9397 & 0.9310 &  0.9002 & 0.8998  \\
      BERT \cite{devlin2018bert} &  0.9305 & 0.9306 & 0.9660 & 0.9569 & 0.9313 & 0.9315  \\
        \bottomrule
    \end{tabular}%
  \label{tab:main_result}%
\end{table*}%
\subsection{Compared Methods}

\noindent\textbf{Dataless} \cite{chang2008importance} is performed based on vector similarity between documents and label surface names using explicit semantic analysis representation. The prediction is the category that yields the maximum cosine similarity. 

\noindent\textbf{Label-Word Similarity} is performed based on the vector similarity between words generated from an MLM and label surface names using the static representation. 
The prediction is the category that yields the maximum cosine similarity. 

\noindent\textbf{Pseudo-CNN} assigns pseudo labels to documents in the training set based on label-word similarity. 
We train a CNN model using pseudo labeled samples in the training set. 
More details are provided in section \ref{sec:implementation}.

\noindent\textbf{Pseudo-BERT} trains BERT \cite{devlin2018bert} {\tt BERT-base-uncased} using the same pseudo labeled data as Pseudo-CNN. More details are provided in section \ref{sec:implementation}.

\noindent\textbf{WeSTClass} \cite{meng2018weakly} first generates pseudo labels for documents which contain user-provided keywords. 
It pre-trains a neural network using pseudo samples as the training set and then performs a self-training process. 

\noindent\textbf{LOTClass} \cite{meng2020text} constructs a category vocabulary for each class, using a pre-trained LM. 
The vocabulary contains words that are relevant to the label name. 
LOTClass fine-tunes an LM via word-level category prediction task, and then performs self-training on unlabeled data to generalize the model.

\noindent\textbf{ConWea} \cite{mekala2020contextualized} leverages contextualized representations of word occurrences and seed word information to automatically distinguish multiple senses of the same word. The contextualized corpus is used to train the classifier and expand seed words iteratively.

\noindent\textbf{X-Class} \cite{wang2021xclass} leverages BERT representations to generate class-oriented document presentations, then generates document-class pairs by clustering, and then fed pairs to a supervised model to train a text classifier. 

\noindent\textbf{CNN}\cite{kim2014convolutional} trains a text CNN using annotated training data in a supervised manner. It is an upper bound of weakly supervised methods.

\noindent\textbf{BERT} fine-tunes BERT {\tt BERT-base-uncased} \cite{devlin2018bert}  using annotated training data. 
It is an upper bound of weakly supervised methods.

\noindent\textbf{WDDC} We use a text CNN\cite{kim2014convolutional} as the document classifier. 
Instead of randomly initializing CNN, we pre-train CNN using Pseudo-CNN.
\noindent\textbf{WDDC-MLM} uses the supervision signals from an MLM while \noindent\textbf{WDDC-Doc} uses the supervision signals from the document itself.


\subsection{Result Analysis}

Table \ref{tab:main_result} shows that our method outperforms weakly supervised baselines by 2\%, 4\%, and 
3\% in AGNews, 20Newsgroup, and UCINews, respectively. 
The gaps between the upper bound CNN and our method are 2\%, 8\%, and 8\% in AGNews, 20Newsgroup, and UCINews, respectively. There are still large performance gaps on 20Newsgroup and UCINews.

Label-Word Similarity and Dataless both use vector similarity for prediction. 
Label-Word Similarity consistently outperforms Dataless, which shows that words generated from an MLM are useful compared with documents. 
The performance of 
Pseudo-BERT is comparable with WeSTClass in AGNews and better than any other baselines in 20Newsgroup and UCINews, which also shows the effectiveness of our pseudo label generation technique. 
In 20Newsgroup, Macro F1 scores are lower than Micro F1 scores in Pseudo-CNN, Pseudo-BERT, and WDDC-MLM methods. 
We found that the number of pseudo labeled data of \textit{sale} category is much lower than other categories.  
So CNN does not have enough pseudo labeled data to learn the \textit{sale} category. The F1 score of \textit{sale} category is lower. 

\begin{table*}[t!]
  \centering
  \caption{Mean and standard deviation of micro and macro F1 scores on 5 independent runs.}
  \begin{tabular}{l|cccc|cccc}
  \toprule
  \multirow{3}{*}{\diagbox{Dataset}{Method}} & \multicolumn{4}{c|}{WDDC} &   \multicolumn{4}{c}{Baselines}  \\
  & \multicolumn{2}{c}{Micro F1} &  \multicolumn{2}{c|}{Macro F1} &   \multicolumn{2}{c}{Micro F1}   &   \multicolumn{2}{c}{Macro F1}\\
   & {~~}Mean{~~} & {~~}Std{~~} & {~~}Mean{~~} & {~~}Std{~~} & {~~}Mean{~~} & {~~}Std{~~} & {~~}Mean{~~} & {~~}Std{~~}\\
  \midrule
AGNews  & 0.8826 & 0.0013 & 0.8825 & 0.0013 & 0.8630 & 0.0038 & 0.8626 & 0.0037\\

20Newsgroup & 0.8570 & 0.0023 & 0.8250 & 0.0033 & 0.8153 & 0.0131 & 0.6896 & 0.0063 \\ 

UCINews & 0.8150 & 0.0012 & 0.8134 &  0.0014 & 0.7824 & 0.0141 & 0.7820 & 0.0148\\
  \bottomrule
  \end{tabular}
  \label{tab:p_value}
\end{table*}

\begin{table*}[ht!]
  \centering
  \caption{Some incorrect predictions in AGNews, 20Newsgroup, and UCINews.}
  \begin{tabular}{l|l|l|l|l}
  \toprule
  Dataset & Text & Prediction & Ground Truth & Signal Words (MLM)\\
  \midrule[1pt]
  AGNews & Microsoft and Palmone today & technology & business & windows, microsoft,  \\
  
  & announced a partnership & & & business, security, \\
  
  & that will likely have a negative  & & &  technology, linux,  \\
  & impact on good technology, & & & privacy \\
  & a well capitalized startup. & & & \\
  
  \midrule
  20News- & For the system, or `family', & computer & science  & software, virus, \\ 
  group & key would appear to be  & & & linux,  encryption , \\ 
  &  cryptographically useless. ... & & & ibm , nsa  \\ 
  & The same key is used for  & & & \\
  & both encryption and decryption.  & & & \\
  \midrule
  UCINews & Paraplegic teenager to kick off & entertainment &  health & football, sport, soccer  \\
  & World Cup thanks to robot suit. & & & cricket, tennis \\
  \bottomrule
  \end{tabular}
  \label{tab:case_study}
\end{table*}

In AGNews and UCINews, WDDC-MLM outperforms WDDC-Doc by 2\% and 3\%, respectively, which shows that signal words from an MLM are more useful than extracted words from a document. 
But in 20Newsgroup, WDDC-Doc outperforms WDDC-MLM by 4\%. 
The possible reason is that some categories in 20Newsgroup are not completely disjoint.  
According to general knowledge, encryption is a field of computer, and computer is a field of science. 
But in 20Newsgroup (refer to Table \ref{tab:stat}), science and encryption belong to one class, and computer belongs to another class. 
MLMs can capture general knowledge from training corpora such as Wikipedia. 
When given a document talking about encryption, an MLM probably generates words about encryption as well as computer. In this circumstance, generated words are misleading while extracted words are clean. We have detailed analysis in section \ref{sec:case_study}.

Table \ref{tab:p_value} shows mean and standard deviation of micro and macro F1 scores of WDDC and best baselines on 5 independent runs. We also conducted t-tests, and p-values are all less than 0.001. We concluded that our method outperforms baselines significantly. Baselines refers to LOTClass, Pseudo-BERT, and Pseudo-BERT on AGNews, 20Newsgroup, and UCINews respectively. 

\subsection{Case Study}
\label{sec:case_study}
\subsubsection{Analysis of Incorrect Predictions}
Table \ref{tab:case_study} shows some incorrect predictions. 
In the first example, some words in the original document such as ``partnership'' and ``startup'' indicate \textit{business} while other words such as ``Microsoft'' and ``technology'' indicate \textit{technology}. 
Signal words generated from an MLM are all related to \textit{technology}. 
In AGNews dataset, there are a number of samples talking about the stock price of technology companies or cooperation between technology companies. 
An MLM inclines to focus on either \textit{technology} or \textit{business} and ignore the other one. 
Although the extraction method can cover all words, the model is likely to be confused when signal words are related to two categories. 
In the second example, an MLM generates words related to \textit{encryption} as well as \textit{computer}. 
Generated words make sense because according to general knowledge, \textit{encryption} is related to \textit{computer}. 
Unfortunately, most of the signal words from an MLM are related to \textit{computer} except one word ``encryption.''
WDDC-MLM is likely to predict it as \textit{computer}. 
Signal words extracted from the document are ``encryption'' and ``key'', which are more likely to guide the model to predict the correct category. 
In the third example, an MLM generates words that are all about sports because the term ``World Cup'' appears in the original document. 
The modifier ``paraplegic'' plays an important role in identifying the true category. 
Both generation and extraction methods fail to capture that. 
\begin{table*}[ht!]
  \centering
   \caption{Top $15$ signal words that have large inner product values with different latent variable vectors respectively on AGNews dataset. Signal words are generated by an MLM. }
  \begin{tabular}{l|l}
  \toprule
  Label & Signal Words\\
  \midrule[1pt]
  Politics & iraq, syria, haiti, israel, murder, baghdad, suicide, \\
  & torture, war, islam, iran, terrorist, afghanistan, religion, terrorism \\
  \midrule
  Sports & injury, racing, baseball, soccer, boxing, player, relegation, \\
  & cricket, quarterback, england, basketball, doping, football, golf, tennis \\
  \midrule
  Business &  profit, market, finance, agriculture, bankruptcy, energy, money, \\
  & growth, price, insurance, recession, airline, oil, risk, inflation \\
  \midrule
  Technology & ipod, genetics, encryption, microsoft, internet, hacking, virus, \\
  & biotechnology, science, copyright, itunes, nasa, evolution, space, astronomy \\
  \bottomrule
  \end{tabular}
  \label{tab:word_weight}
\end{table*}
\subsubsection{Analysis of Word Distribution Learner}
The word distribution learner aims to estimate the probability of a signal word $w_r$ given a possible value of category $c$, i.e., $p(w_r|c)$. 
A good word distribution learner should assign a high probability to category-indicated words, so that by maximizing Eq. (\ref{eq:objective_one}), a large value of $p(w_r|c)$ leads to a large value of $q(c|x)$, which means if a document contains indicative words to category $c$, it possibly belongs to category $c$. 
Table \ref{tab:word_weight} shows top $15$ signal words that have large inner product values with different latent variable vectors respectively on AGNews dataset.  
As shown in Table \ref{tab:word_weight}, the selected words are category-indicated. 
For example, in the \textit{politics} category, all words are about terrorism, war, and places where wars broke out, which are relevant to the \textit{politics} topic. 
The word distribution learner can be consider as a category-indicated keywords expansion module. 


\subsection{Implementation}
\label{sec:implementation}
We use the BERT ({\tt bert-base-uncased}) model to obtain supervision signals in AGNews and 20Newsgroup. 
We use the BERT ( {\tt bert-base-cased}) to obtain supervision signals in UCINews which contains many acronyms such as WHO and PTSD. 
We select top $20$ predictions as supervision signals three datasets. To remove non-discriminated words, we set the threshold $t$ to $2$ in three datasets.

In the pseudo label generation process, we set the threshold $\gamma$ to $0.6$, $0.75$, and $0.55$ in AGNews, 20Newsgroup, and UCINews, respectively. Those pseudo labeled training data are used in Pseudo-CNN and Pseudo-BERT. 
A higher $\gamma$ may result in more accurate pseudo labels. 
But we need to balance the size of pseudo labeled data because it will shrink when $\gamma$ increases.

To train WDDC, in each batch, we randomly select $5$ signal words among all signal words of a document. 
The number of negative samples in the approximated word distribution learner is set to $10$. 
For Pseudo-CNN, CNN, and WDDC methods, the CNN architectures are the same. 
Four different filter sizes $\{2, 3, 4, 5\}$ are applied. 
A max-pooling layer is applied to each convolutional layer, and each convolutional layer has $100$ filters. 
The maximum length of input in the CNN is set to $64$, $128$, and $64$ in AGNews, 20Newsgroup, and UCINews,  respectively. 
The input in the CNN is contextualized embeddings generated by BERT ({\tt bert-base-uncased}).

For WeSTClass, we use a CNN as the document classifier because it empirically outperforms LSTM in WeSTClass. The CNN architecture we used here is the same as the one described in their paper. 
We try our best to find good keywords and tune hyper-parameters for WeSTClass and LOTClass.  
For all methods, we tune hyper-parameters on development sets. 


\section{Conclusion}
To solve the weakly supervised classification task, we propose to query a masked language model with cloze style  prompts to obtain supervision signals. 
We design a prompt which combines the document itself and ``this article is talking about {\tt[MASK]}.''
The predictions for the ``{\tt [MASK]}'' token are considered as supervision signals because they summarize the content of documents.
We propose a latent variable model (WDDC) to learn word distributions given pre-defined categories and a neural document classifier simultaneously without using any annotated data.
Evaluation on three datasets shows that our method can outperform weakly supervised learning baselines. 
\section*{Acknowledgements}
The authors of this paper were partially supported by the NSFC Fund (U20B2053) from the NSFC of China, the RIF (R6020-19 and R6021-20) and the GRF (16211520) from RGC of Hong Kong, the MHKJFS (MHP/001/19) from ITC of Hong Kong and the National Key R\&D Program of China (2019YFE0198200) with special thanks to Hong Kong Mediation and Arbitration Centre (HKMAAC) and California University, School of Business Law \& Technology (CUSBLT), and the Jiangsu Province Science and Technology Collaboration Fund (BZ2021065).
We also thank the anonymous reviewers for their valuable comments and suggestions that help improve the quality of this manuscript.
\bibliographystyle{acl_natbib}
\bibliography{custom}

\begin{thebibliography}{40}
\expandafter\ifx\csname natexlab\endcsname\relax\def\natexlab#1{#1}\fi

\bibitem[{Caragea et~al.(2011)Caragea, McNeese, Jaiswal, Traylor, Kim, Mitra,
  Wu, Tapia, Giles, Jansen, and Yen}]{caragea2011classifying}
Cornelia Caragea, Nathan~J. McNeese, Anuj~R. Jaiswal, Greg Traylor, Hyun{-}Woo
  Kim, Prasenjit Mitra, Dinghao Wu, Andrea~H. Tapia, C.~Lee Giles, Bernard~J.
  Jansen, and John Yen. 2011.
\newblock Classifying text messages for the haiti earthquake.
\newblock In \emph{Proceedings of ISCRAM}.

\bibitem[{Chang et~al.(2008)Chang, Ratinov, Roth, and
  Srikumar}]{chang2008importance}
Mingwei Chang, Lev~Arie Ratinov, Dan Roth, and Vivek Srikumar. 2008.
\newblock Importance of semantic rep- resentation: dataless classification.
\newblock In \emph{Proceedings of AAAI}, pages 830--835.

\bibitem[{Chen et~al.(2018)Chen, Xiong, Yan, and Wang}]{chen2018variational}
Wenhu Chen, Wenhan Xiong, Xifeng Yan, and William Wang. 2018.
\newblock Variational knowledge graph reasoning.
\newblock In \emph{Proceedings of NAACL-HLT}, pages 1823--1832.

\bibitem[{Devlin et~al.(2019)Devlin, Chang, Lee, and
  Toutanova}]{devlin2018bert}
Jacob Devlin, Ming-Wei Chang, Kenton Lee, and Kristina Toutanova. 2019.
\newblock Bert: Pre-training of deep bidirectional transformers for language
  understanding.
\newblock In \emph{Proceedings of NAACL-HLT}, pages 4171--4186.

\bibitem[{Dua and Graff(2017)}]{dua2017uci}
Dheeru Dua and Casey Graff. 2017.
\newblock \href {http://archive.ics.uci.edu/ml} {{UCI} machine learning
  repository}.

\bibitem[{Gabrilovich et~al.(2007)Gabrilovich, Markovitch
  et~al.}]{gabrilovich2007computing}
Evgeniy Gabrilovich, Shaul Markovitch, et~al. 2007.
\newblock Computing semantic relatedness using wikipedia-based explicit
  semantic analysis.
\newblock In \emph{Proceedings of IJCAI}, pages 1606--1611.

\bibitem[{Jensen et~al.(1906)}]{jensen1906fonctions}
Johan Ludwig William~Valdemar Jensen et~al. 1906.
\newblock Sur les fonctions convexes et les in{\'e}galit{\'e}s entre les
  valeurs moyennes.
\newblock \emph{Acta mathematica}, 30:175--193.

\bibitem[{Johnson and Zhang(2017)}]{johnson2017deep}
Rie Johnson and Tong Zhang. 2017.
\newblock Deep pyramid convolutional neural networks for text categorization.
\newblock In \emph{Proceedings of ACL}, pages 562--570.

\bibitem[{Kim(2014)}]{kim2014convolutional}
Yoon Kim. 2014.
\newblock Convolutional neural networks for sentence classification.
\newblock In \emph{Proceedings of EMNLP}, pages 1746--1751.

\bibitem[{Kingma and Welling(2014)}]{KingmaW13}
Diederik~P Kingma and Max Welling. 2014.
\newblock Auto-encoding variational bayes.
\newblock In \emph{Proceedings of ICLR}.

\bibitem[{Kristina et~al.(2003)Kristina, Dan, Christopher, and
  Singer}]{toutanova2003feature}
Toutanova Kristina, Klein Dan, Manning Christopher, and Yoram Singer. 2003.
\newblock Feature-rich part-of-speech tagging with a cyclic dependency network.
\newblock In \emph{Proceedings of NAACL-HLT}, pages 252--259.

\bibitem[{Lang(1995)}]{lang1995newsweeder}
Ken Lang. 1995.
\newblock Newsweeder: Learning to filter netnews.
\newblock In \emph{Proceedings of ICML}, pages 331--339.

\bibitem[{Liang et~al.(2019)Liang, Liu, Zhang, and Song}]{liang2019relation}
Yan Liang, Xin Liu, Jianwen Zhang, and Yangqiu Song. 2019.
\newblock Relation discovery with out-of-relation knowledge base as
  supervision.
\newblock In \emph{Proceedings of NAACL-HLT}, pages 3280--3290.

\bibitem[{Liu et~al.(2019)Liu, Zhang, Fan, Fu, Li, Wu, and
  Lam}]{liu2019reconstructing}
Han Liu, Xiaotong Zhang, Lu~Fan, Xuandi Fu, Qimai Li, Xiao{-}Ming Wu, and
  Albert Y.~S. Lam. 2019.
\newblock Reconstructing capsule networks for zero-shot intent classification.
\newblock In \emph{Proceedings of EMNLP}, pages 4798--4808.

\bibitem[{Marcheggiani and Titov(2016)}]{marcheggiani2016discrete}
Diego Marcheggiani and Ivan Titov. 2016.
\newblock Discrete-state variational autoencoders for joint discovery and
  factorization of relations.
\newblock \emph{Transactions of the Association for Computational Linguistics},
  4:231--244.

\bibitem[{Mekala and Shang(2020)}]{mekala2020contextualized}
Dheeraj Mekala and Jingbo Shang. 2020.
\newblock Contextualized weak supervision for text classification.
\newblock In \emph{Proceedings of ACL}, pages 323--333.

\bibitem[{Meng et~al.(2018)Meng, Shen, Zhang, , and Han}]{meng2018weakly}
Yu~Meng, Jiaming Shen, Chao Zhang, , and Jiawei Han. 2018.
\newblock Weakly-supervised neural text classification.
\newblock In \emph{Proceedings of CIKM}, pages 983--992.

\bibitem[{Meng et~al.(2019)Meng, Shen, Zhang, , and Han}]{meng2019weakly}
Yu~Meng, Jiaming Shen, Chao Zhang, , and Jiawei Han. 2019.
\newblock Weakly-supervised hierarchical text classification.
\newblock In \emph{Proceedings of AAAI}, pages 6826--6833.

\bibitem[{Meng et~al.(2020)Meng, Zhang, Huang, Xiong, Ji, Zhang, and
  Han}]{meng2020text}
Yu~Meng, Yunyi Zhang, Jiaxin Huang, Chenyan Xiong, Heng Ji, Chao Zhang, and
  Jiawei Han. 2020.
\newblock Text classification using label names only: A language model
  self-training approach.
\newblock In \emph{Proceedings of EMNLP}, pages 9006--9017.

\bibitem[{Mikolov et~al.(2013)Mikolov, Sutskever, Chen, Corrado, and
  Dean}]{mikolov2013distributed}
Tomas Mikolov, Ilya Sutskever, Kai Chen, Greg~S Corrado, and Jeff Dean. 2013.
\newblock Distributed representations of words and phrases and their
  compositionality.
\newblock In \emph{Proceedings of NeurIPS}, pages 3111--3119.

\bibitem[{Petroni et~al.(2020)Petroni, Lewis, Piktus, Rockt{\"a}schel, Wu,
  Miller, and Riedel}]{petroni2020context}
Fabio Petroni, Patrick Lewis, Aleksandra Piktus, Tim Rockt{\"a}schel, Yuxiang
  Wu, Alexander~H Miller, and Sebastian Riedel. 2020.
\newblock How context affects language models' factual predictions.
\newblock In \emph{Proceedings of AKBC}.

\bibitem[{Petroni et~al.(2019)Petroni, Rockt{\"a}schel, Lewis, Bakhtin, Wu,
  Miller, and Riedel}]{petroni2019language}
Fabio Petroni, Tim Rockt{\"a}schel, Patrick Lewis, Anton Bakhtin, Yuxiang Wu,
  Alexander~H Miller, and Sebastian Riedel. 2019.
\newblock Language models as knowledge bases?
\newblock In \emph{Proceedings of EMNLP}, pages 2463--2473.

\bibitem[{Radford et~al.(2019)Radford, Wu, Child, Luan, Amodei, and
  Sutskever}]{radford2019language}
Alec Radford, Jeff Wu, Rewon Child, David Luan, Dario Amodei, and Ilya
  Sutskever. 2019.
\newblock Language models are unsupervised multitask learners.
\newblock \emph{OpenAI blog}.

\bibitem[{Rezende et~al.(2014)Rezende, Mohamed, and
  Wierstra}]{rezende2014stochastic}
Danilo~Jimenez Rezende, Shakir Mohamed, and Daan Wierstra. 2014.
\newblock Stochastic backpropagation and approximate inference in deep
  generative models.
\newblock In \emph{Proceedings of ICML}, pages 1278--1286.

\bibitem[{Rios and Kavuluru(2018)}]{kavuluru2018riloff}
Anthony Rios and Ramakanth Kavuluru. 2018.
\newblock Few-shot and zero-shot multi-label learning for structured label
  spaces.
\newblock In \emph{Proceedings of EMNLP}, pages 3132--3142.

\bibitem[{Roitblat et~al.(2010)Roitblat, Kershaw, and
  Oot}]{roitblat2010document}
Herbert~L. Roitblat, Anne Kershaw, and Patrick Oot. 2010.
\newblock Document categorization in legal electronic discovery: computer
  classification vs. manual review.
\newblock \emph{Journal of the Association for Information Science and
  Technology}, 61(1):70--80.

\bibitem[{Schick and Sch{\"u}tze(2021)}]{schick2021exploiting}
Timo Schick and Hinrich Sch{\"u}tze. 2021.
\newblock Exploiting cloze questions for few shot text classification and
  natural language inference.
\newblock In \emph{Proceedings of EACL}, pages 255--269.

\bibitem[{Schick and Schütze(2021)}]{schick2021small}
Timo Schick and Hinrich Schütze. 2021.
\newblock It's not just size that matters: Small language models are also
  few-shot learners.
\newblock In \emph{Proceedings of NAACL-HLT}, pages 2339--2352.

\bibitem[{Shin et~al.(2020)Shin, Razeghi, Logan~IV, Wallace, and
  Singh}]{shin2020autoprompt}
Taylor Shin, Yasaman Razeghi, Robert~L Logan~IV, Eric Wallace, and Sameer
  Singh. 2020.
\newblock Autoprompt: Eliciting knowledge from language models with
  automatically generated prompts.
\newblock In \emph{Proceedings of EMNLP}, pages 4222--4235.

\bibitem[{Song and Roth(2014)}]{song2014on}
Yangqiu Song and Dan Roth. 2014.
\newblock On dataless hierarchical text classification.
\newblock In \emph{Proceedings of AAAI}, pages 1579--1585.

\bibitem[{{\v{S}}uster et~al.(2016){\v{S}}uster, Titov, and van
  Noord}]{vsuster2016bilingual}
Simon {\v{S}}uster, Ivan Titov, and Gertjan van Noord. 2016.
\newblock Bilingual learning of multi-sense embeddings with discrete
  autoencoders.
\newblock In \emph{Proceedings of NAACL-HLT}, pages 1346--1356.

\bibitem[{Titov and Khoddam(2015)}]{titov2014unsupervised}
Ivan Titov and Ehsan Khoddam. 2015.
\newblock Unsupervised induction of semantic roles within a
  reconstruction-error minimization framework.
\newblock In \emph{Proceedings of NAACL-HLT}, pages 1--10.

\bibitem[{Wang et~al.(2021)Wang, Mekala, and Shang}]{wang2021xclass}
Zihan Wang, Dheeraj Mekala, and Jingbo Shang. 2021.
\newblock X-class: Text classification with extremely weak supervision.
\newblock In \emph{Proceedings of NAACL-HLT}, pages 3043--3053.

\bibitem[{Xia et~al.(2018)Xia, Zhang, Yan, Chang, and Yu}]{xia2018zero}
Congying Xia, Chenwei Zhang, Xiaohui Yan, Yi~Chang, and Philip~S. Yu. 2018.
\newblock Zero-shot user intent detection via capsule neural networks.
\newblock In \emph{Proceedings of EMNLP}, pages 3090--3099.

\bibitem[{Zeng et~al.(2019)Zeng, Zhou, Liu, and Song}]{zeng2019variational}
Ziqian Zeng, Wenxuan Zhou, Xin Liu, and Yangqiu Song. 2019.
\newblock A variational approach to weakly supervised document-level
  multi-aspect sentiment classification.
\newblock In \emph{Proceedings of NAACL-HLT}, pages 386--396.

\bibitem[{Zhang et~al.(2019)Zhang, Lertvittayakumjorn, and
  Guo}]{zhang2019integraing}
Jingqing Zhang, Piyawat Lertvittayakumjorn, and Yike Guo. 2019.
\newblock Integrating semantic knowledge to tackle zero-shot text
  classification.
\newblock In \emph{Proceedings of NAACL-HLT}, pages 1031--1040.

\bibitem[{Zhang et~al.(2022)Zhang, Yu, Shetty, Song, and
  Zhang}]{zhang2022prboost}
Rongzhi Zhang, Yue Yu, Pranav Shetty, Le~Song, and Chao Zhang. 2022.
\newblock Prboost: Prompt-based rule discovery and boosting for interactive
  weakly-supervised learning.
\newblock In \emph{Proceedings of ACL}.

\bibitem[{Zhang et~al.(2015)Zhang, Zhao, and LeCun}]{zhang2015character}
Xiang Zhang, Junbo~Jake Zhao, and Yann LeCun. 2015.
\newblock Character-level convolutional networks for text classification.
\newblock In \emph{Proceedings of NeurIPS}, pages 649--657.

\bibitem[{Zhang et~al.(2018)Zhang, Dai, Kozareva, Smola, and
  Song}]{zhang2018variational}
Yuyu Zhang, Hanjun Dai, Zornitsa Kozareva, Alexander~J Smola, and Le~Song.
  2018.
\newblock Variational reasoning for question answering with knowledge graph.
\newblock In \emph{Proceedings of AAAI}, pages 6069--6076.

\bibitem[{Zhou et~al.(2016)Zhou, Qi, Zheng, Xu, Bao, and Xu}]{zhou2016text}
Peng Zhou, Zhenyu Qi, Suncong Zheng, Jiaming Xu, Hongyun Bao, and Bo~Xu. 2016.
\newblock Text classification improved by integrating bidirectional lstm with
  two-dimensional max pooling.
\newblock In \emph{Proceedings of COLING}, pages 3485--3495.

\end{thebibliography}
\end{document}